\documentclass{article}
\usepackage{float}

\usepackage{PRIMEarxiv}
\usepackage{caption}
\usepackage{subcaption}
\usepackage[flushleft]{threeparttable}
\usepackage{tabularx}

\usepackage[utf8]{inputenc} 
\usepackage[T1]{fontenc}    
\usepackage{hyperref}       
\usepackage{url}            
\usepackage{booktabs}       
\usepackage{amsfonts}       
\usepackage{nicefrac}       
\usepackage{microtype}      
\usepackage{lipsum}
\usepackage{graphicx}
\graphicspath{{media/}}     

  \usepackage[dvipsnames]{xcolor}

\title{Increasing Textual Context Size Boosts Medical Image-Text Matching}

\author{
  Idan Glassberg \\
  The Hebrew University of Jerusalem \\
  \texttt{idan.glassberg@mail.huji.ac.il} \\
   \And
  Tom Hope \\
  The Hebrew University of Jerusalem \\
  \texttt{tom.hope@mail.huji.ac.il} \\
}

\begin{document}
\maketitle
\fancyhead[R]{}

\begin{abstract}
This short technical report demonstrates a simple technique that yields state of the art results in medical image-text matching tasks. We analyze the use of OpenAI's CLIP\cite{clip}, a general image-text matching model, and observe that CLIP's limited textual input size has negative impact on downstream performance in the medical domain where encoding longer textual contexts is often required. We thus train and release ClipMD, which is trained with a simple sliding window technique to encode textual captions. ClipMD was tested on two medical image-text datasets and compared with other image-text matching models. The results show that ClipMD outperforms other models on both datasets by a large margin. We make our code and pretrained model publicly available\footnote{\url{https://github.cs.huji.ac.il/tomhope-lab/ClipMD}.}.
\end{abstract}


\section{Introduction}
Pretrained image-text matching models, such as OpenAI's CLIP \cite{clip}, use natural language processing (NLP) approaches to find semantic relations between images and textual descriptions. This emerging technology has seen rapid adoption in the general domain, and increasing interest in the medical domain \cite{BioViL,PubMedCLIP} where medical imaging data often includes images paired with textual descriptions. For example, MIMIC-CXR\cite{MIMIC-CXR} is a dataset that consists of chest radiographs along with free-text radiology reports. This dataset paved the way for works like BioViL \cite{BioViL} which used the images and the captions provided in the dataset to train an image-text matching model for chest X-Rays and chest related diseases. ROCO \cite{ROCO} is a dataset containing radiology images from publications available in the PubMed biomedical paper repository. ROCO includes several medical imaging modalities beyond X-Ray, such as CT,  Ultrasound and MRI. PubMedCLIP \cite{PubMedCLIP} was trained on ROCO as part of a medical visual question answering model.

Models like PubMedCLIP are based on fine-tuning of pretrained general image-text matching models \cite{clip}. However, in this technical report we observe that currently these powerful pretrained models are typically trained on images with \textit{short text captions}, and thus their text input size is much smaller than the textual information in medical image-text datasets. This forces existing medical image-text matching models to use only a fraction of the caption given to each image in their input data, since long medical descriptions cannot fit within their small input size.

In this brief technical report, our goal is to study whether simple remedies can be applied to make better use of the power of image-text models in the medical domain. We observe that by simply modifying CLIP's fine-tuning to encode the input text captions with a sliding window technique, we are able to dramatically boost image-text matching performance. On the ROCO dataset, we obtain improvements of 10-20\% over state of the art models \cite{PubMedCLIP, MedICaT}.

\section{Method}
\subsection{Model}
Our simple fine-tuning modification and the resulting model is named ClipMD. ClipMD is a fine-tuned model, with the base pretrained model being OpenAI's CLIP, a leading general-domain image-text matching model. Specifically, we used Hugging Face's Clip ViT32 model\footnote{https://huggingface.co/openai/clip-vit-base-patch32}. This model uses a ViT (Visual Transformer) to encode images and a text encoder with an input size of 77 tokens, the same input size as all other pretrained CLIP models available.

\begin{figure}[h]
  \centering
  \includegraphics[scale=0.6]{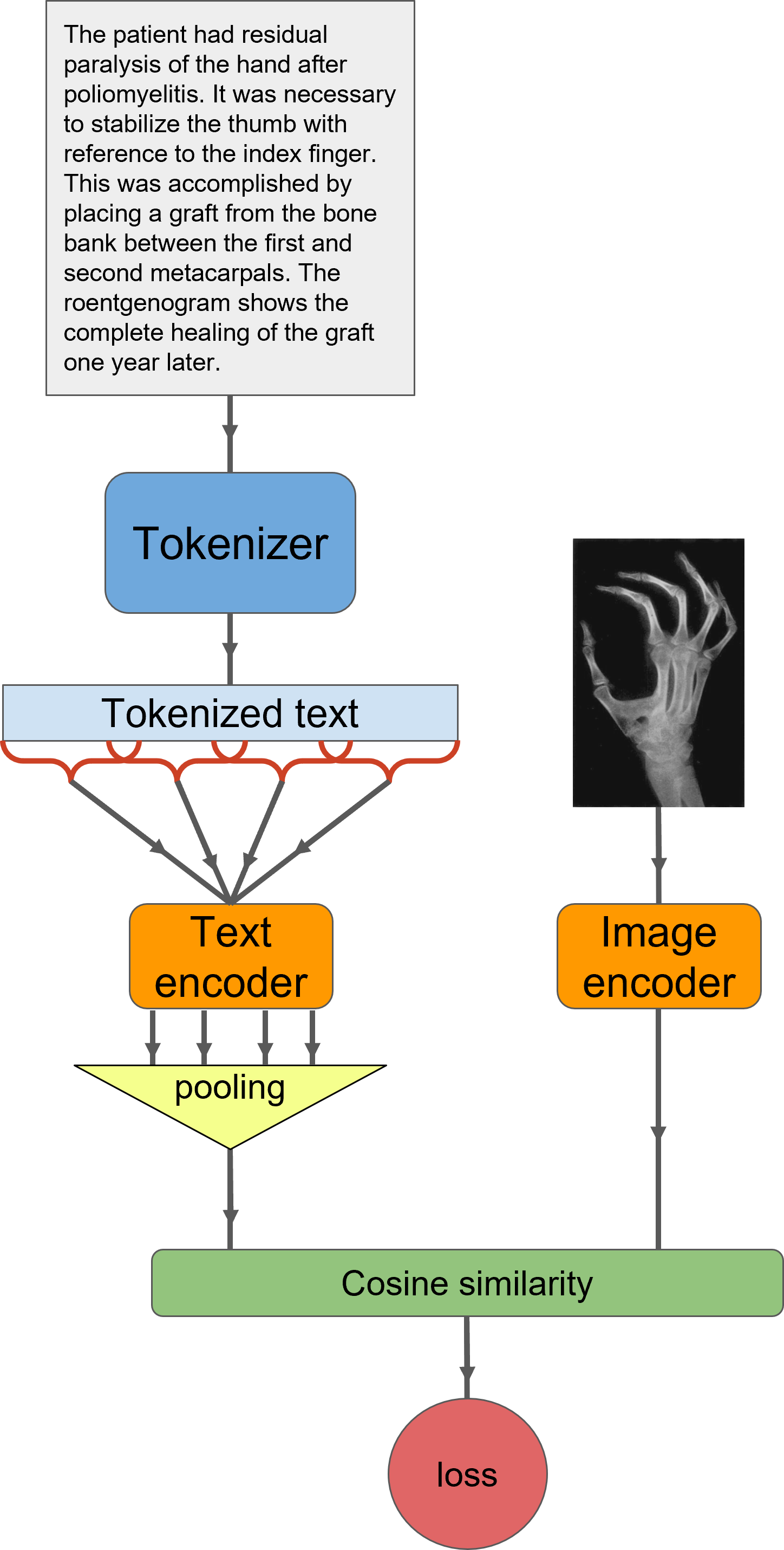}
  \caption{An overview of the adapted CLIP fine-tuning for incorporating long captions in medical image-text matching.}
    \label{fig:fig1}
\end{figure}

Unlike other existing work in this area, after tokenizing the text captions we feed them into our model's encoder using a sliding overlapping window and taking the mean of the encoder's output (see Figure \ref{fig:fig1}). We tried 3×3, 5×5, and 7×7 average pooling over the encoder's output, but 1×1 pooling worked best in this case. This commonly applied technique in other models and tasks\cite{Long-Short}, empirically leads to surprisingly large gains in the medical domain, spurring our publication of this report to inform practitioners and future research in this space. 

\subsection{Datasets}
We tested ClipMD on two different medical image-text datasets:
\begin{enumerate}
    \item The ROCO\cite{ROCO} dataset consists of around 82K non-compound radiology images (X-rays, CTs, MRIs...) and their captions from the paper they were taken from. The data is split into a training set of size 65K images, a validation of size 8K images, and a test set of size 8K images. The dataset also includes the UMLS\cite{UMLS} concept unique IDs and semantic types for each image.
    \item The MedICaT\cite{MedICaT} dataset consists of 217K images generated by radiology, histology, and other visual scoping procedures. The dataset includes both compound and non-compound images. We split the data into a training set, validation set and test set with the same ratios as the ROCO dataset.
\end{enumerate}
\subsection{Experimental setup}
We fine-tuned ClipMD on both datasets using the Pytorch\cite{pytorch} framework, with Adam optimization (learning rate of $10^{-6}$) for 10 epochs and batch size of 50. We trained and ran our experiments on a GPU cluster owned by The Hebrew University of Jerusalem. Since our experiments included random sampling, we repeated the experiments 5 times and report the average Recall$@$K scores. We compare our results on the ROCO dataset only with the results of models with publicly available pretrained weights tuned on the ROCO dataset. For the MedICaT dataset, we compare to the base CLIP model before fine-tuning as we have not been able to find image-text matching models that were trained on this dataset.

\section{Results}
We randomly sampled 2000 image/caption pairs from each dataset.  We compare our results on ROCO with four other models that were fine-tuned on the ROCO dataset. Three of them are PubMedCLIP\cite{PubMedCLIP} models with various image encoders and the fourth model is a SciBRET\cite{SciBERT} based image-text matching model proposed in the MedICaT\cite{MedICaT} paper (see Table \ref{tab:table1}). Our model performed 10-20\% better than the other models we used in our experiment. This large performance boost must come as a result of the sliding window method we propose, since it was the only factor that differentiated our model from the PubMedCLIP model that uses ViT32 as an image encoder.
\begin{table}[H]
\centering
\begin{threeparttable}
 \caption{Recall$@$K comparison on ROCO.}
  \begin{tabular}{lllll}
    \toprule
    Name     & R@1	& R@5	& R@10	& R@20 \\
    \cmidrule(r){1-5}
    PubMedCLIP/RN50 & $7.1_{0.11}$ & $21_{0.21}$ & 32$_{0.24}$ & $45_{0.15}$    \\
    PubMedCLIP/RN50x4 &  $7.7_{0.12}$ & $23_{0.25}$ & $34_{0.55}$ & $48_{0.43}$ \\
    PubMedCLIP/ViT32 & $8.5_{0.17}$ & $26_{0.32}$ & $38_{0.22}$ & $53_{0.45}$   \\
    MedICaT-SciBERT & $7.6_{0.59}$ & $26_{1.6}$ & $41_{1.6}$ & $58_{1.3}$   \\
    \cmidrule(r){1-5}
    ClipMD &  \boldmath{$17_{0.35}$} & \boldmath{$40_{0.44}$} & \boldmath{$54_{0.37}$} & \boldmath{$68_{0.49}$}           \\
    \bottomrule
  \end{tabular}
  \begin{tablenotes}
      \small
      \item Average results over n = 5 random seeds. The error (in subscript) is the standard error $\sigma/\sqrt{n}$, where $\sigma$ is the standard deviation over random seeds.
    \end{tablenotes}
  \label{tab:table1}
\end{threeparttable}
\end{table}

The second dataset we used in our experiment was the MedICaT dataset. Since there aren't any image-text matching models that were trained or fine-tuned specifically on MedICaT, we compared our results with the base CLIP image-text matching model before the fine-tuning process (see Table \ref{tab:table2}). Our model's Recall$@$K scores were better than CLIP's by between 25\% to 60\% at K $=20$. CLIP wasn't fine-tuned on the dataset and its training data wasn't medically focused, which partially explains its low recall at k scores.  
\begin{table}[H]
\centering
\begin{threeparttable}
 \caption{Recall$@$K comparison on MedICaT.}
  
  \begin{tabular}{p{3cm}llll}
    \toprule
    Name     & R@1	& R@5	& R@10	& R@20 \\
    \cmidrule(r){1-5}
    CLIP/ViT32        & $3.2_{0.19}$ & $9_{0.19}$ & $13_{0.11}$ & $19_{0.16}$ \\
    \cmidrule(r){1-5}
    ClipMD &  \boldmath{$29_{0.43}$} & \boldmath{$57_{0.44}$} & \boldmath{$69_{0.44}$} & \boldmath{$79_{0.58}$}           \\
    \bottomrule
  \end{tabular}
  \label{tab:table2}
  \begin{tablenotes}
      \small
      \item Results show mean percent accuracy with error in subscript over n = 5 random seeds. The error is the standard error $\sigma/\sqrt{n}$, where $\sigma$ is the standard deviation over random seeds.
    \end{tablenotes}
    \end{threeparttable}
\end{table}

\section{Preliminary Error Analysis}

\begin{figure}[h]
  \centering
  \caption{\textcolor{green}{Correct} and \textcolor{red}{incorrect} examples of image-text matching using ClipMD.}
    \label{fig:fig3}
  \begin{subfigure}[h]{0.3\textwidth}
    \includegraphics[width=\textwidth]{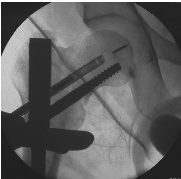}
  \textbf{Ground truth}: Fluoroscopic view showing insertion of scope into screw tunnel.\\
  \textbf{Top prediction}: \textcolor{green}{Fluoroscopic view showing insertion of scope into screw tunnel}.
  \end{subfigure}
  \hfill
  \begin{subfigure}[h]{0.3\textwidth}
  \includegraphics[width=\textwidth]{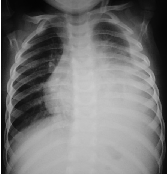}
  \textbf{Ground truth}: Radiography de thorax montrant une pleurésie grande abundance.

  \textbf{Top prediction}: \textcolor{green}{Radiography de thorax montrant une pleurésie grande abundance}.
  \end{subfigure}
  \hfill
  \begin{subfigure}[h]{0.3\textwidth}
  \includegraphics[width=\textwidth]{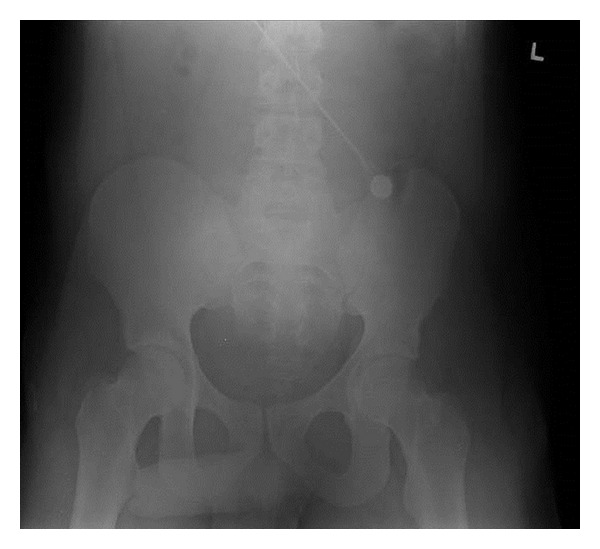}
  \textbf{Ground truth}: Pelvic X-ray did not reveal any fracture or radiopaque foreign body.

  \textbf{Top prediction}: \textcolor{green}{Pelvic X-ray did not reveal any fracture or radiopaque foreign body}.
  \end{subfigure}
\end{figure}

We briefly provide examples of correct and incorrect captions our model matched to images from the ROCO dataset. As demonstrated in Figure \ref{fig:fig3}, incorrect captions often included the correct part of the body that is shown in the image and the kind of imaging technology used, while the specific disease or abnormality present in the images was missed. For the most part these false matches occurred for diseases that are underrepresented in the training data or for captions that reference a part of their original paper that is not included (such as the bottom-right example that refers to a future LVAD implantation that is not shown in the image).

\begin{figure}[h]
  \centering
  \begin{subfigure}[h]{0.3\textwidth}
    \includegraphics[width=\textwidth]{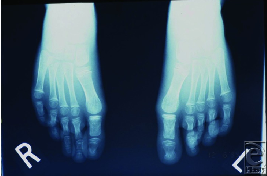}
  \textbf{Ground truth}: Radiograph of feet showing osteohypertrophy of the first metatarsal.

  \textbf{Top prediction}: \textcolor{red}{Skiagram of feet shwoing lytic changes in involved bones}.
  \end{subfigure}
  \hfill
  \begin{subfigure}[h]{0.3\textwidth}
  \includegraphics[width=\textwidth]{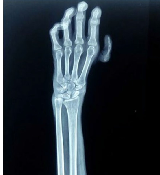}
  \textbf{Ground truth}: X-ray anteroposterior (AP) view of hand showing absent first metacarpal.

  \textbf{Top prediction}: \textcolor{red}{Anteroposterior X-ray of a patient affected by TAR syndrome showing complete aplasia of the radius and a triphalangeal thumb}.
  \end{subfigure}
  \hfill 
  \begin{subfigure}[h]{0.3\textwidth}
  \includegraphics[width=\textwidth]{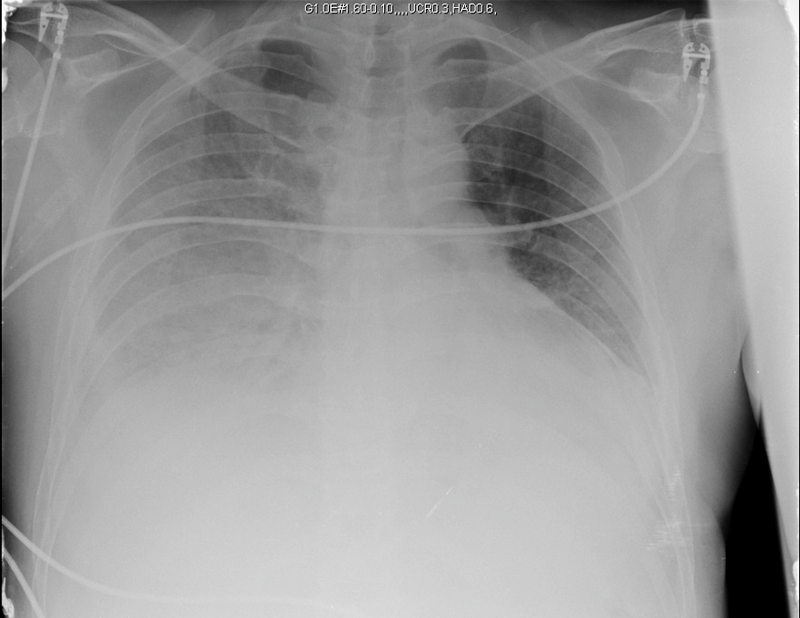}
  \textbf{Ground truth}: Chest X-ray before LVAD implantation.

  \textbf{Top prediction}: \textcolor{red}{Chest radiograph showing left-sided pleural effusion shortly after admission}.
  \end{subfigure}
\end{figure}


\section{Conclusion}
Our simple sliding window ``patch'' gives fine-tuned pretrained image-text matching models the ability to encode the entire context of a candidate medical caption before matching it with an image. We publicly release ClipMD that utilizes the sliding window. ClipMD achieves state of the art performance on both of the datasets we used in our experiments. This demonstrates that providing image-text matching models with the full context of the input captions makes a dramatic difference in performance without the need to change and retrain a new text encoder.

\bibliographystyle{unsrt}  
\bibliography{references}

\end{document}